%% file: main.tex
\documentclass[10pt,twocolumn,letterpaper]{article}

\usepackage[pagenumbers]{iccv} 

\input{preamble}

\definecolor{iccvblue}{rgb}{0.21,0.49,0.74}
\usepackage[pagebackref,breaklinks,colorlinks,allcolors=iccvblue]{hyperref}
\usepackage{booktabs}   
\usepackage{multirow}   
\usepackage{graphicx}
\usepackage{amsmath}

\usepackage{cuted}

\def\paperID{14467} 
\def\confName{ICCV}
\def\confYear{2025}

\title{DISTIL: Data-Free Inversion of Suspicious Trojan Inputs via Latent Diffusion}

\author{\textbf{Hossein Mirzaei, Zeinab Taghavi, Sepehr Rezaee, Masoud Hadi, Moein Madadi,} \\\textbf{Mackenzie W. Mathis } \\
École Polytechnique Fédérale de Lausanne (EPFL) \\\texttt{hossein.mirzaeisadeghlou@epfl.ch, mackenzie.mathis@epfl.ch
} \\
}

\begin{document}
\maketitle
\input{sec/0_abstract}    
\input{sec/1_intro}

\input{sec/3_Related_Works}
\input{sec/4_Method}

\input{sec/5_Experiment}
\input{sec/6_Ablation_Study}
\input{sec/7_Conclusion}

{
    \small
    \bibliographystyle{unsrtnat}
    \bibliography{main}
}

\newpage
\input{sec/8_Appendix}

\end{document}

%% file: preamble.tex
\usepackage{algorithm}
\usepackage[noend]{algpseudocode}
\usepackage{amsmath}
\usepackage{amssymb}
\usepackage{float}

%% file: sec/0_abstract.tex
\begin{abstract}
Deep neural networks have demonstrated remarkable success across numerous tasks, yet they remain vulnerable to Trojan (backdoor) attacks, raising serious concerns about their safety in real-world mission-critical applications. A common countermeasure is trigger inversion -- reconstructing malicious ``shortcut'' patterns (triggers) inserted by an adversary during training. 
Current trigger-inversion methods typically search the full pixel space under specific assumptions but offer no assurances that the estimated trigger is more than an adversarial perturbation that flips the model output.
Here, we propose a data-free, zero-shot trigger-inversion strategy that restricts the search space while avoiding strong assumptions on trigger appearance. Specifically, we incorporate a diffusion-based generator guided by the target classifier; through iterative generation, we produce candidate triggers that align with the internal representations the model relies on for malicious behavior. Empirical evaluations, both quantitative and qualitative, show that our approach reconstructs triggers that effectively distinguish clean versus Trojaned models. DISTIL surpasses alternative methods by high margins, achieving up to \textbf{7.1\%} higher accuracy on the BackdoorBench dataset and a \textbf{9.4\%} improvement on trojaned object detection model scanning, offering a promising new direction for reliable backdoor defense \textbf{without} reliance on extensive data or strong prior assumptions about triggers. The code is available at \url{https://github.com/AdaptiveMotorControlLab/DISTIL}.

\end{abstract}

%% file: sec/1_intro.tex
\input{sec/Fig_1}

\section{Introduction} \label{sec:intro}

As artificial intelligence continues to rapidly evolve, detecting Trojan attacks in models has become a critical challenge. Trojan attacks, which insert malicious trigger patterns into training data, allow models to function normally on clean inputs but mispredict inputs containing these triggers~\cite{gu2019badnets,blended}. Recently, these attacks have grown more potent by leveraging sophisticated label mapping techniques and enhancing stealthiness through dynamic or invisible triggers~\cite{sig,inputaware,nguyen2021wanet,color,bpp,ssba}. Trojan attacks pose a significant threat to safety-critical computer vision applications, such as autonomous driving and object detection, where undetected triggers could lead to catastrophic failures~\cite{gao2020backdoor,saha2020hidden,li2022backdoorlearningsurvey,han2022physical,zhang2022towards,li2023backdoor,zhaosurvey}.

In response to these attacks, researchers have developed a variety of defense strategies to detect and mitigate Trojaned models \cite{li2021anti,liu2022backdoor,chen2022effective,huang2022backdoor,zhao2024adversarially}. Among these, methods that reverse engineer triggers (RET) have emerged as a critical post-training defense mechanism \cite{NC,umd}. RET methods estimate trigger patterns based on the model behavior, often analyzing output confidence levels. Early RET methods typically optimized for a small patch in the image that acted as a proxy for the trigger. More recent approaches relax prior assumptions about trigger characteristics by integrating feature-space information or employing alternative regularization strategies \cite{wang2019neural,hu2022trigger,shen2021backdoor,tao2022better,wang2022rethinking,wang2023unicorn}. Notably, all these techniques assume access to clean training data for performing pixel-space optimization. Once reconstructed, the estimated trigger can be used to scan Trojaned models, mitigate attacks, and predict target classes~\cite{xu2023towards}.

Despite their success, existing RET methods can conflate actual Trojan triggers with adversarial perturbations, leading to false positives in Trojan scanning \cite{santurkar2019image,sun2023single,mazeika2023Trojan}.
High-dimensional pixel-space optimization often leads to adversarial noise rather than true triggers, compromising the effectiveness of existing RET methods~\cite{szegedy2013intriguing,Su_2019,weng2020trade,niu2024towards}. This limitation results in noisy or less interpretable triggers and increases false positives when scanning benign models. Adapting these methods to other tasks such as  object detection is challenging due to spatial variability, multi-output structures, and post-processing complexities~\cite{shen2023django,mazeika2023Trojan}. Moreover, reliance on clean data limits real-world applicability, as datasets are often inaccessible~\cite{sun2023single,MMBD}.

To overcome these challenges, we introduce \textbf{DISTIL}: \textbf{D}ata-free \textbf{I}nversion of \textbf{S}uspicious \textbf{T}rojan \textbf{I}nputs via \textbf{L}atent diffusion, a novel method that estimates interpretable and discriminative triggers \textit{without} requiring any clean training samples. Our approach shifts the optimization process from the pixel space to a pre-trained guided diffusion model's latent space, thereby reducing the risk of finding purely adversarial artifacts and increasing the likelihood of uncovering legitimate trigger patterns.

Our key insight is that even the most sophisticated Trojaned models are distinctly biased toward specific transferable shortcut patterns. By guiding a diffusion backward process with under test model gradients, we generate synthetic patterns that closely mimic these triggers, thereby enabling a range of defense capabilities. Specifically, we use the recovered patterns to scan Trojaned models, identify target label, and mitigate attacks. Notably, pre-trained guided diffusion models have already been trained to follow gradient-based guidance signals. Their pretraining inherently equips them to easily adapt to new guidance, such as that provided by our test models. This inherent adaptability allows DISTIL to seamlessly extend its capabilities to different scenarios, including the scanning of Trojaned object detection or classifier models.

%% file: sec/Fig_1.tex
\begin{figure*}[ht]
    \centering
    \includegraphics[width=.99\textwidth]{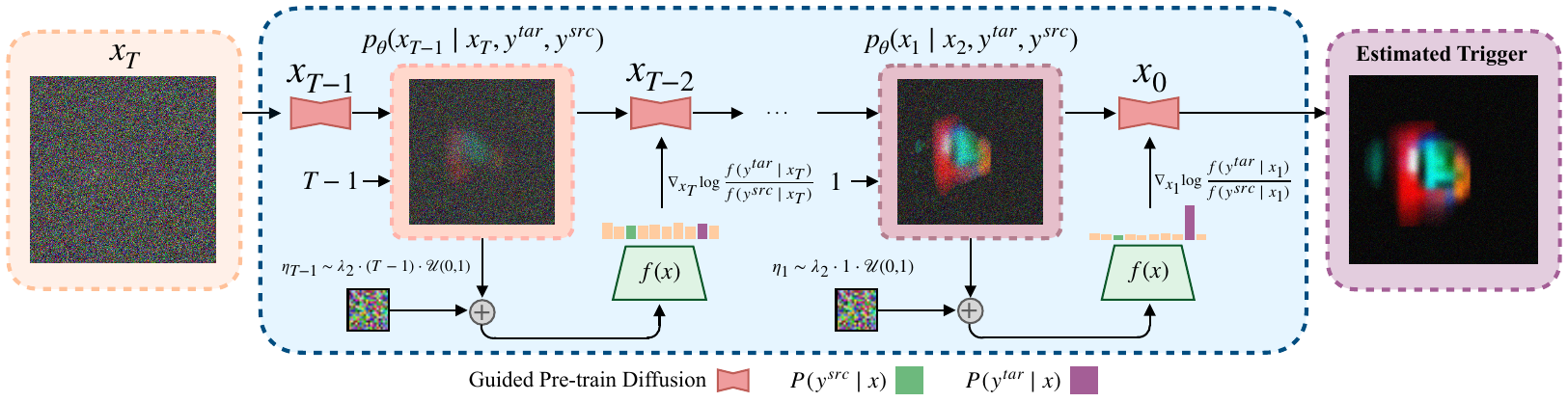}
    \vspace{-2.5mm}
\caption{\textbf{Method overview.}  
DISTIL inverts Trojan triggers without any clean data by steering a pre‑trained, classifier‑guided diffusion model. The process begins with pure Gaussian noise $x_T \sim \mathcal{N}(0,I)$ and iteratively refines it toward a trigger‐like pattern using the classifier’s gradients for the chosen objective. At every diffusion step $t$, we inject additional uniform noise $\eta_t$ before re‑evaluating the gradients, which regularises the search and discourages convergence to adversarial perturbations. Because the diffusion backbone was pre‑trained to follow gradient guidance, it can faithfully track these signals in latent space and reveal genuine shortcut patterns. Finally, exploiting the fact that such shortcuts transfer more strongly in Trojaned networks, DISTIL reliably distinguishes compromised models from clean ones.}

    \label{fig:trigger_inversion_pipeline}
\end{figure*}

%% file: sec/3_Related_Works.tex
\input{sec/Fig_2}

\vspace{-3pt}
\section{Related Works}

\textbf{Trojan Attacks.}
Trojan attacks have grown increasingly sophisticated, utilizing diverse strategies for manipulating label mappings and employing stealthy, dynamic triggers. Label manipulation techniques range from simpler all-to-one mappings to more complex one-to-one, one-to-all, and all-to-all mappings. Early methods like BadNet \cite{gu2019badnets} introduced visible, static triggers, while newer approaches such as SIG \cite{sig} and WaNet \cite{nguyen2021wanet} focus on stealth through imperceptible. Dynamic attacks such as InputAware \cite{inputaware} and BppAttack \cite{bpp} create sample-specific triggers, complicating defense, and highlight the critical need for advanced protective measures.

\noindent\textbf{RET for Trojan Attack Defense.} Trigger reconstruction serves as a defense method against Trojan attacks by using the estimated trigger strategy for various tasks. NC \cite{wang2019neural} serves as a baseline for reverse engineering defenses by generating small static perturbations in pixel space. Subsequent efforts, including FeatureRE~\cite{wang2022rethinking} and Pixel~\cite{BetterTrigger}, introduced feature space constraints and improved optimization techniques to enhance trigger fidelity. K-Arm \cite{kARM} employed multi-arm bandits to explore potential attack classes more efficiently. THTP \cite{hu2022trigger} leveraged topological priors to refine trigger patterns and better localize suspicious regions. Meanwhile, UMD~\cite{umd} addressed the challenge of varying label mapping attacks by jointly inferring arbitrary source-target mappings without relying on prior knowledge of target labels. UNICORN~\cite{wang2023unicorn} unified trigger inversion across diverse spaces (pixel, signal, feature, numerical) by employing input space transformations and formulating the inversion as an optimization problem with multiple constraints. BTI-DBF \cite{xu2024towards} decouples benign and Trojan features during optimization by employing a dual-branch architecture to enhance trigger estimation. In response to the tendency to extract adversarial perturbations rather than triggers, SmoothInv \cite{sun2023single} aimed to robustify under the test classifier by applying randomized smoothing before pixel space optimization, thereby limiting its applicability to patch-based attacks.

%% file: sec/Fig_2.tex
\begin{figure*}[t]
  \centering
  \includegraphics[width=\linewidth]{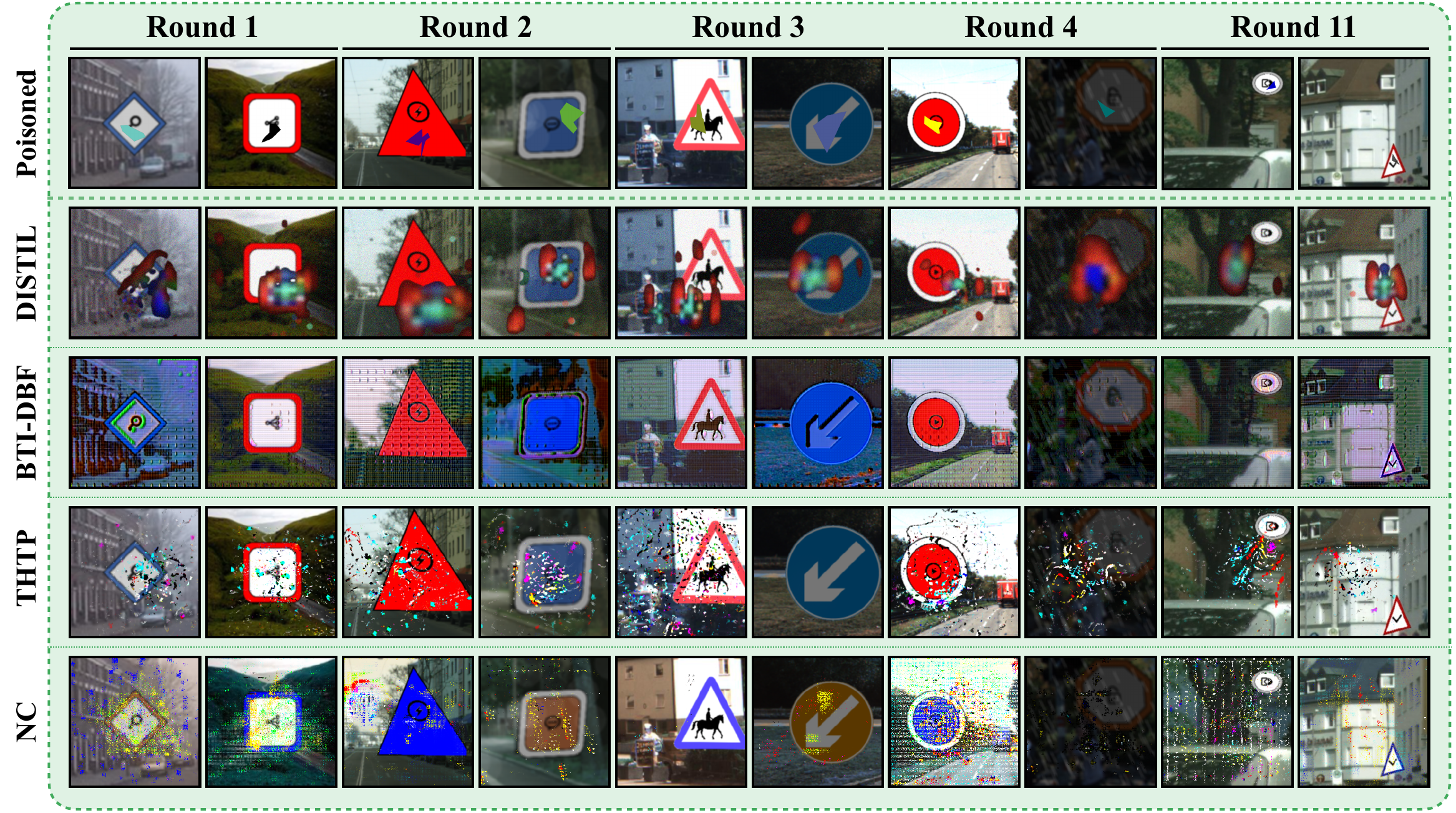}
  \caption{\textbf{Visual comparison of trigger‑inversion methods.}
  Estimated triggers produced by prior RET baselines (columns) across multiple TrojAI rounds often resemble adversarial noise rather than true triggers.  
  By constraining the search to the latent space of a guided diffusion model and regularising with uniform‑noise augmentation, \textsc{DISTIL} instead uncovers shortcut patterns that closely match the genuine triggers.}
  \label{fig:trigger_comparison}
\end{figure*}

%% file: sec/4_Method.tex
\section{Method}

\textbf{Motivation.} 
The goal of Trojan scanning is to identify signatures that distinguish Trojaned models from their clean counterparts \cite{su2024model, MMBD}. Our central hypothesis is that shortcut patterns learned by Trojaned models demonstrate significantly greater transferability. This arises because a Trojaned network is explicitly trained to link a specific trigger to a target class, thereby establishing strong spurious correlations that induce misclassifications whenever the trigger appears. Although clean models may also latch onto natural shortcuts during training, they remain far less sensitive to these specific patterns \cite{he2023mitigatingbackdoorpoisoningattacks, li2024shortcutsnowhereexploringmultitrigger}.

Our first aim, therefore, is to extract and estimate these shortcut patterns, then define the Trojan signature by measuring how the model’s predictions change after embedding the estimated shortcut into clean inputs. For a Trojaned model, the shift toward the attack’s target label should be large; for a clean model, it should be much smaller.

Directly optimizing pixel values to recover these shortcuts is problematic, however. The optimization can collapse onto adversarial perturbations rather than genuine triggers, and because both benign and Trojaned models are susceptible to adversarial noise, the resulting patterns cannot serve as a discriminative signature. Worse still, if the Trojan was implanted using adversarial training, pixel‑space optimization may yield a pattern that fools the clean model more than the Trojaned one, leading to false positives.

To avoid these pitfalls we move the search into the latent space of a pretrained image‑diffusion generator.  
At each denoising step we steer the generator with the gradient of an objective that increases the probability of the Trojan’s target class while decreasing that of the corresponding source class.  
Because the generator’s manifold is constrained to realistic images, the search space is far narrower than raw pixel space and is less prone to degenerate, adversarial artifacts.  
We further inject small uniform noise into the classifier input at every reverse‑diffusion step; this acts as a regulariser, discouraging brittle solutions and nudging the optimisation toward robust, transferable shortcut patterns.

The resulting pattern serves as an interpretable signature that we employ for (i) Trojan scanning, (ii) mitigation via fine‑tuning, and (iii) prediction of the attack’s target class. Subsequent sections provide full details of our DISTIL method: Figure \ref{fig:trigger_inversion_pipeline} gives a high‑level overview, while Figure \ref{fig:trigger_comparison} compares the shortcuts estimated by DISTIL with those recovered by prior approaches.

\textbf{Threat Model.}  
We consider an adversary who injects a small set of poisoned samples during training. Each poisoned sample contains a trigger \(T\) stamped onto a source-class image \(x\) and is mislabeled as a target-class \(y^\ast\). Consequently, the model learns to associate \(T\) with \(y^\ast\). At inference time, the backdoored model behaves normally on clean inputs but classifies any trigger-containing input as \(y^\ast\), thus enabling targeted misclassifications while evading detection under standard testing. In general, a backdoor attack against a classifier with \(N\) classes, \(f : \mathcal{X} \to \mathcal{Y}\), is defined by a trigger embedding function \(\delta : \mathcal{X} \to \mathcal{X}\) and a set \(A \subset \mathcal{Y} \times \mathcal{Y}\) of backdoor class pairs.

\textbf{Diffusion Guided by Classifier for RET.} To reconstruct  Trojan triggers, we employ a pretrained guided diffusion model~\cite{nichol2021glide} to follow classifier under test gradients. Our objective is to simultaneously increase the likelihood of a designated target class \(y^{\text{tar}}\) and decrease the likelihood of a source class \(y^{\text{src}}\). In doing so, we encourage the diffusion model to reveal shortcut patterns learned by the Trojaned classifier. To further ensure that the diffusion process does not converge to mere adversarial perturbations, we incorporate an additional safeguard. At each reverse diffusion step, random \emph{uniform} noise is injected into the classifier input. This uniform noise forces the diffusion model to discover genuine trigger patterns, patterns to which the Trojaned classifier is inherently vulnerable, rather than simply finding adversarial artifacts. 

  Formally, we modify the mean of the reverse process as follows:\begin{equation}
\begin{aligned}
    \tilde{\mu}_{\theta}(x_t, t, y^{\text{tar}}, y^{\text{src}}) &= 
    \mu_{\theta}(x_t, t) + \\
    &\Sigma_{\theta}(x_t, t) \nabla_{x_t} \log   \frac{f(y^{\text{tar}} \mid x_t)}{f(y^{\text{src}} \mid x_t)} + \lambda_1\cdot\eta_t,
\end{aligned}
\label{eq:mu_tilde}
\end{equation}
Where \( \tilde{\mu}_{\theta}(x_t, t, y^{\text{tar}}, y^{\text{src}}) \) is the modified mean of the reverse process at time step \( t \) for a given input \( x_t \) with corresponding target and source labels \( y^{\text{tar}} \) and \( y^{\text{src}} \), respectively, \( \mu_{\theta}(x_t, t) \) is the original mean of the reverse process, \( \Sigma_{\theta}(x_t, t) \) is the covariance matrix for the reverse process, \( \nabla_{x_t} \log \frac{f(y^{\text{tar}} \mid x_t)}{f(y^{\text{src}} \mid x_t)} \) is the gradient of the classifier logits corresponding to target class minus source class for the input \( x_t \). The gradient term of the \ref{eq:mu_tilde}, is computed with respect to the input $x_t$ at each diffusion step, using the classifier's logits to guide the generation toward patterns that shift predictions from the source class to the target class.  \( \eta_t \sim \mathcal{U}(0,1) \) represents the uniform noise term and \( \lambda_1 \) is a hyperparameter controlling its intensity. This uniform noise injection acts as a regularizer, disrupting brittle adversarial perturbations that are sensitive to small changes, thereby encouraging the diffusion process to converge on robust, trigger-like patterns inherent to the Trojaned model.

The next sample is then drawn from the distribution:
\begin{equation}
    p_{\theta}(x_{t-1} \mid x_t, y^{\text{tar}}, y^{\text{src}}) = \mathcal{N}\left(\tilde{\mu}_{\theta}(x_t, t, y^{\text{tar}}, y^{\text{src}}), \Sigma_{\theta}(x_t, t)\right),
\end{equation}

\noindent where \( \mathcal{N}(\mu, \Sigma) \) denotes a normal distribution with mean \( \mu \) and covariance \( \Sigma \).

When clean source-class data denoted as \( \mathcal{X}^{\text{src}} \) are available, DISTIL optionally enhances its reconstruction through hybrid conditioning. We modify the gradient term in Equation~\ref{eq:mu_tilde} as follows:
\begin{equation} \label{Equation:hybrid}
    \nabla_{x_t} \log \frac{f (y^{\text{tar}} \mid \mathcal{X}^{\text{src}} \oplus x_t)}{f (y^{\text{src}} \mid \mathcal{X}^{\text{src}} \oplus x_t)},
\end{equation}

We then use the final generated image \( x_0 \) as the trigger corresponding to the pair \( (y^{\text{src}}, y^{\text{tar}}) \), denoted as \( \delta^{\text{tar}}_{\text{src}} \), if the probability assigned by the classifier exceeds a threshold \( \lambda_2 \) (e.g., 0.95), meaning:
\begin{equation}
    \text{softmax}\left[f(\delta^{\text{tar}}_{\text{src}})\right]_{y^{\text{tar}}} \geq \lambda_2,
\end{equation}
where \( \text{softmax}[f(x)] \) is the softmax function applied to the classifier's output for input \( x \), which provides the predicted class probabilities, \( \left[\cdot\right]_{y^{\text{tar}}} \) extracts the predicted probability for the target class \( y^{\text{tar}} \), and \( \lambda_2 \) is a threshold parameter that determines the minimum probability required to consider the trigger as valid.
Otherwise, the generation process is repeated until such a trigger for the considered source and target labels is achieved. We limit repeated generation by a maximum upper bound, and these hyperparameters have been discussed in the experimental details.

\textbf{Trojan Detection and Mitigation.} For a classifier under test, we define a \emph{signature}, a scalar score, to quantify the likelihood that the classifier has been compromised by a Trojan attack. This score captures the \emph{transferability} of an extracted shortcut: it measures how strongly an estimated trigger shifts the classifier’s prediction from a source class \(y^{\text{src}}\) to a target class \(y^{\text{tar}}\).   In particular, the trigger is embedded into held-out images from the source class to assess its effect. Specifically, for a given trigger \( \delta^{\text{tar}}_{\text{src}} \), our signature, defined as the trigger strength (i.e., transferability), is evaluated as follows
\begin{equation}
\begin{split}
    \text{Score}(\delta^{\text{tar}}_{\text{src}}; f) = \mathbb{E}_{x' \sim \mathcal{X}'^{\text{src}}} \Big[ 
    & \mathrm{softmax}\Big(f\big(x' + \delta^{\text{tar}}_{\text{src}}\big)\Big)_{y^{\text{tar}}} \\
    & - \mathrm{softmax}\Big(f\big(x' + \delta^{\text{tar}}_{\text{src}}\big)\Big)_{y^{\text{src}}} \Big],
\end{split}
\label{eq:trigger_score}
\end{equation}
The overall Trojan score of the classifier is then defined as the maximum score over all possible target classes and their corresponding source-class triggers:
\begin{equation}
    \text{Score}(f) = \max_{y^{\text{tar}}} \max_{y^{\text{src}} \neq y^{\text{tar}}} \text{Score}(\delta^{\text{tar}}_{\text{src}}; f).
\label{eq:overall_score}
\end{equation}

The trigger achieving the maximum score in Equation~\ref{eq:overall_score} target is predicted as the target class of the Trojaned classifier. Notably, this is meaningful when the attack targets a single class; for label mapping strategies such as all-to-all, there is no specific target class.

For mitigation, we create a dataset by injecting the triggers into clean images from their corresponding source classes while preserving the correct labels. The classifier is then finetuned on this dataset, which helps the model focus on authentic image features rather than being misled by shortcut cues introduced by the triggers. We note that using 1\% of clean data for Trojan scanning and mitigation is common in the literature \cite{bsurvey}, and we adopt this setting when running experiments with baselines.

\input{Tables/table_1}

\textbf{Fast DISTIL.}  
Exhaustively scanning every $(y^{\text{src}},y^{\text{tar}})$ pair scales quadratically with the number of classes, $O(K^{2})$, and quickly becomes impractical. Fast DISTIL lowers this cost to $O(K)$ without sacrificing accuracy. For each prospective target class $y^{\text{tar}}$ we identify a single, maximally distant source class
\begin{equation}
y^{\text{src}} = \arg\!\min_{\,y\neq y^{\text{tar}}}
      \; \cos\bigl(\phi(y),\,\phi(y^{\text{tar}})\bigr),
      \label{eq:cosine}
\end{equation}
where $\phi(\cdot)$ denotes the mean feature vector in the network’s penultimate layer and $\cos(\cdot,\cdot)$ is cosine similarity. Selecting the farthest class leverages the intuition that a trigger capable of shifting predictions from the  most dissimilar class to $y^{\text{tar}}$ must be exploiting an especially strong, model‐specific shortcut; if such a shortcut exists, it will be revealed here before anywhere else. In practice this heuristic slashes computation by an order of magnitude while maintaining the same detection power, as confirmed in our ablation (Setup D of Table \ref{table:ablation_study}).

\textbf{Adapting DISTIL to Object Detection.}  
To adapt DISTIL for scanning object‑detection networks, we augment the guidance term in Equation~\ref{eq:mu_tilde} with an additional gradient that encourages a spatial shift in the detector’s predictions. We add
$\nabla_{x_t}\!\log P\bigl(\text{bbox}\!\rightarrow\!\text{corner}\,\big|\,x_t\bigr),$
where \(P(\text{bbox}\!\rightarrow\!\text{corner}\mid x_t)\) is the model’s probability that the centre of each predicted bounding‑box falls inside a pre‑selected corner region.  The combined gradient therefore simultaneously steers the classifier’s output from the source class toward the target class and drags bounding boxes toward the chosen corner, which is sampled uniformly at random for every input data. At evaluation time the estimated trigger is added to the entire image.  The detector’s Trojan score is computed as the sum of the classification shift in Equation~\ref{eq:trigger_score} and the mean displacement of ground‑truth bounding boxes measured on held‑out data.  A large score indicates a strong, transferable shortcut and thus a high probability that the detector is Trojaned (see Figure~\ref{fig:fig3}).

%% file: Tables/table_1.tex
\begin{table*}[p]
\centering
\caption{Comparison of scanning performance between the proposed DISTIL method and alternative approaches. Tables~1-a and 1-b summarize performance on Trojaned \textbf{classifier} models, while Table~1-c reports results for Trojaned \textbf{object detection} models, all reported in terms of accuracy. The best results in each column are highlighted in bold.}
\begin{subtable}[t]{\textwidth}
\caption{Comparison of scanning performance between DISTIL and alternative methods on the BackdoorBench dataset, covering various attack scenarios.}
\label{table:backdoorbench_model_detection}

\resizebox{\textwidth}{!}{%
\begin{tabular}{ll c c c c c c c c c c c c}
\specialrule{1.5pt}{\aboverulesep}{\belowrulesep}
    \multirow{3}{*}{\textbf{Dataset}} & \multirow{3}{*}{\textbf{Attack}} & \multicolumn{12}{c}{\textbf{Method}} \\
    \cmidrule(lr){3-14}
    & & \multirow{2}{*}{\textbf{NC}} 
      & \multirow{2}{*}{\textbf{ABS}} 
      & \multirow{2}{*}{\textbf{Pixel}} 
      & \multirow{2}{*}{\textbf{\small THTP}} 
      & \multirow{2}{*}{\textbf{\footnotesize SmoothInv}} 
      & \multirow{2}{*}{\textbf{\small Unicorn}} 
      & \multirow{2}{*}{\textbf{\footnotesize BTI-DBF}} 
      & \multirow{2}{*}{\textbf{K-Arm}} 
      & \multirow{2}{*}{\textbf{\small MM-BD}} 
      & \multirow{2}{*}{\textbf{\small TRODO}} 
      & \multirow{2}{*}{\textbf{UMD}} 
      & \textbf{\small DISTIL} \\
    & & & & & & & & & & & & & \textbf{\textit{\small (Ours)}} \\
\specialrule{1.5pt}{\aboverulesep}{\belowrulesep}
\multirow{11}{*}{\rotatebox[origin=c]{90}{\textbf{CIFAR-10}}}       & \textbf{BadNets}   & 76.4 & 73.5 & 74.0 & 75.6 & 86.3 & 82.9 & 84.0 & 75.7 & 81.3 & 86.2 & 76.8 & \cellcolor{blue!15}\textbf{94.9} \\
      & \textbf{Blended}   & 65.2 & 70.8 & 67.6 & 65.2 & 84.9 & 78.2 & 85.7 & 73.5 & 74.6 & 85.0 & 69.4 & \cellcolor{blue!15}\textbf{93.4} \\
      & \textbf{BPP}       & 62.5 & 64.0 & 58.9 & 60.9 & 75.5 & 75.4 & 76.4 & 55.4 & 72.2 & 83.9 & 63.9 & \cellcolor{blue!15}\textbf{88.7} \\
      & \textbf{inputaware}& 58.1 & 53.9 & 56.2 & 49.7 & 69.7 & 73.1 & 79.2 & 58.1 & 65.9 & 71.7 & 58.2 & \cellcolor{blue!15}\textbf{93.2} \\
      & \textbf{LC}        & 62.8 & 56.6 & 51.8 & 54.2 & 74.4 & 67.5 & \cellcolor{blue!15}\textbf{90.5} & 59.6 & 80.4 & 81.2 & 59.0 & 89.5 \\
      & \textbf{LF}        & 68.3 & 61.7 & 53.4 & 62.5 & 81.6 & 74.0 & 83.3 & 63.2 & 79.3 & 78.0 & 65.1 & \cellcolor{blue!15}\textbf{91.0} \\
      & \textbf{LIRA}      & 54.9 & 58.3 & 46.1 & 65.4 & 70.8 & 66.9 & 80.0 & 54.9 & 81.8 & 82.5 & 57.4 & \cellcolor{blue!15}\textbf{90.6} \\
      & \textbf{SIG}       & 52.0 & 56.5 & 54.6 & 46.8 & 67.3 & 65.6 & 81.9 & 63.8 & 79.5 & 84.8 & 58.3 & \cellcolor{blue!15}\textbf{92.8} \\
      & \textbf{SSBA}      & 66.1 & 47.0 & 61.2 & 52.0 & 79.6 & 73.4 & 72.6 & 57.3 & 78.1 & 81.2 & 62.6 & \cellcolor{blue!15}\textbf{90.3} \\
      & \textbf{TrojanNN}  & 52.5 & 49.2 & 45.0 & 58.3 & 63.0 & 59.1 & 83.7 & 82.7 & 75.0 & 76.4 & 56.2 & \cellcolor{blue!15}\textbf{86.1} \\
      & \textbf{WaNet}     & 63.7 & 57.4 & 52.5 & 54.1 & 68.9 & 65.3 & \cellcolor{blue!15}\textbf{86.8} & 56.1 & 72.7 & 80.0 & 61.7 & 84.4 \\
    \midrule

\multirow{11}{*}{\rotatebox[origin=c]{90}{\textbf{GTSRB}}}       & \textbf{BadNets}    & 75.6 & 67.2 & 70.2 & 68.7 & 86.7 & 83.4 & 85.4 & 76.3 & 81.8 & 87.6 & 75.3 & \cellcolor{blue!15}\textbf{92.1} \\
      & \textbf{Blended}    & 64.8 & 70.1 & 63.9 & 67.1 & 85.0 & 78.1 & 86.8 & 73.1 & 75.4 & 84.3 & 67.6 & \cellcolor{blue!15}\textbf{91.4} \\
      & \textbf{BPP}        & 62.0 & 59.8 & 54.2 & 56.3 & 75.9 & 75.5 & 85.5 & 53.7 & 72.7 & \cellcolor{blue!15}\textbf{89.1} & 61.9 & 86.9 \\
      & \textbf{inputaware} & 51.9 & 66.7 & 58.8 & 47.6 & 69.8 & 73.2 & 79.1 & 57.0 & 66.0 & 64.5 & 57.4 & \cellcolor{blue!15}\textbf{91.3} \\
      & \textbf{LC}         & 57.3 & 48.0 & 61.4 & 54.9 & 75.2 & 67.8 & \cellcolor{blue!15}\textbf{91.3} & 61.6 & 74.5 & 85.2 & 56.2 & 89.8 \\
      & \textbf{LF}         & 61.5 & 53.6 & 50.5 & 62.5 & 81.6 & 74.3 & 81.6 & 57.2 & 79.4 & 80.5 & 63.7 & \cellcolor{blue!15}\textbf{90.5} \\
      & \textbf{LIRA}       & 48.2 & 59.4 & 52.1 & 65.4 & 70.3 & 67.0 & 76.5 & 52.8 & 72.9 & 76.4 & 53.1 & \cellcolor{blue!15}\textbf{88.2} \\
      & \textbf{SIG}        & 52.0 & 56.3 & 54.6 & 46.8 & 67.5 & 65.7 & 80.0 & 62.3 & 79.6 & 67.5 & 56.3 & \cellcolor{blue!15}\textbf{90.9} \\
      & \textbf{SSBA}       & 66.4 & 47.9 & 61.2 & 52.0 & 79.2 & 73.5 & 73.9 & 55.6 & 78.2 & 72.3 & 61.7 & \cellcolor{blue!15}\textbf{85.6} \\
      & \textbf{TrojanNN}   & 52.7 & 49.2 & 45.0 & 58.9 & 63.8 & 59.2 & 82.5 & \cellcolor{blue!15}\textbf{83.9} & 75.1 & 85.5 & 52.1 & 82.7 \\
      & \textbf{WaNet}      & 63.3 & 57.4 & 52.3 & 54.1 & 69.0 & 65.4 & \cellcolor{blue!15}\textbf{88.2} & 54.4 & 72.8 & 71.9 & 60.7 & 86.0 \\
    \midrule

\multirow{11}{*}{\rotatebox[origin=c]{90}{\textbf{Tiny ImageNet}}}  
 
      & \textbf{BadNets}     & 76.7 & 73.8 & 64.1 & 67.8 & 84.9 & 82.3 & 79.0 & 68.6 & 78.6 & 75.0 & 73.1 & \cellcolor{blue!15}\textbf{91.5} \\
      & \textbf{Blended}     & 52.0 & 65.1 & 60.4 & 66.2 & 83.2 & 77.7 & 84.5 & 65.0 & 73.1 & 76.4 & 64.5 & \cellcolor{blue!15}\textbf{89.1} \\
      & \textbf{BPP}         & 64.5 & 57.3 & 46.9 & 54.5 & 73.6 & 68.4 & 82.3 & 54.3 & 72.5 & 74.2 & 59.9 & \cellcolor{blue!15}\textbf{83.7} \\
      & \textbf{inputaware}  & 58.2 & 60.0 & 52.2 & 42.3 & 66.4 & 66.2 & 77.7 & 55.7 & 64.9 & 70.0 & 56.7 & \cellcolor{blue!15}\textbf{90.2} \\
      & \textbf{LC}          & 56.6 & 42.8 & 40.8 & 51.9 & 74.0 & 64.5 & 89.9 & 56.4 & 78.3 & 81.5 & 53.4 & \cellcolor{blue!15}\textbf{86.4} \\
      & \textbf{LF}          & 45.9 & 53.6 & 52.3 & 60.4 & 80.5 & 73.8 & 79.0 & 58.2 & 78.8 & 83.9 & 61.6 & \cellcolor{blue!15}\textbf{87.9} \\
      & \textbf{LIRA}        & 57.3 & 51.5 & 48.0 & 63.6 & 68.8 & 64.6 & 73.4 & 50.6 & 80.5 & 79.3 & 50.2 & \cellcolor{blue!15}\textbf{84.0} \\
      & \textbf{SIG}         & 54.8 & 58.7 & 46.5 & 41.1 & 65.3 & 63.2 & 77.1 & 59.9 & 78.2 & 65.8 & 55.0 & \cellcolor{blue!15}\textbf{89.3} \\
      & \textbf{SSBA}        & 52.2 & 46.4 & 36.1 & 50.3 & 77.6 & 72.9 & 70.8 & 52.7 & 77.4 & 72.1 & 59.3 & \cellcolor{blue!15}\textbf{83.9} \\
      & \textbf{TrojanNN}    & 38.7 & 51.9 & 55.4 & 56.9 & 60.5 & 58.3 & 75.2 & 79.1 & 73.9 & 82.7 & 49.8 & \cellcolor{blue!15}\textbf{81.1} \\
      & \textbf{WaNet}       & 46.4 & 56.0 & 51.7 & 53.0 & 67.1 & 61.8 & 80.6 & 53.4 & 70.0 & 76.5 & 59.5 & \cellcolor{blue!15}\textbf{81.6} \\
\specialrule{1.5pt}{\aboverulesep}{\belowrulesep}
\end{tabular}
}
\end{subtable}


\begin{subtable}[t]{\textwidth}
\caption{Comparison of scanning performance for Trojaned classifier models across multiple rounds of the TrojAI benchmark under diverse attack scenarios.}
\label{table:trojai_model_detection}
\resizebox{\textwidth}{!}{%
\begin{tabular}{l c c c c c c c c c c c c}
\specialrule{1.5pt}{\aboverulesep}{\belowrulesep}
    \multirow{3}{*}{\textbf{Dataset}} & \multicolumn{12}{c}{\textbf{Method}} \\
    \cmidrule(lr){2-13}
    & \multirow{1}{*}{\textbf{NC}}
      & \multirow{1}{*}{\textbf{ABS}}
      & \multirow{1}{*}{\textbf{Pixel}}
      & \multirow{1}{*}{\textbf{\small THTP}}
      & \multirow{1}{*}{\textbf{\footnotesize SmoothInv}}
      & \multirow{1}{*}{\textbf{\small Unicorn}}
      & \multirow{1}{*}{\textbf{\footnotesize BTI-DBF}}
      & \multirow{1}{*}{\textbf{K-Arm}}
      & \multirow{1}{*}{\textbf{\small MM-BD}}
      & \multirow{1}{*}{\textbf{\small TRODO}}
      & \multirow{1}{*}{\textbf{UMD}}
      & \textbf{\small DISTIL} \\
\specialrule{1.5pt}{\aboverulesep}{\belowrulesep}
    \textbf{Round 0}  & 75.1 & 70.3 & 76.6 & 74.7 &78.2   & 72.4 & 82.5 & \cellcolor{blue!15}\textbf{91.3} & 80.5 &  86.2&  80.4  & 83.1 \\
    \textbf{Round 1}  & 72.1 & 68.5 & 71.4 &65.1  & 75.0 & 66.3 & 79.1 & \cellcolor{blue!15} \textbf{90.0}  & 71.5  & 85.7& 79.2 & 82.9 \\
    \textbf{Round 2}  & 63.0 & 61.2 & 58.8 & 62.6 & 67.5  & 58.4 & 68.6 & 76.4 & 55.8  & 78.1& 75.2 & \cellcolor{blue!15} \textbf{79.5} \\
    \textbf{Round 3}  & 61.4 & 57.6 & 52.1 & 61.7 & 65.8 & 56.2 & 64.2 & \cellcolor{blue!15} \textbf{82.0}& 52.6 & 77.2& 61.3  & 78.4 \\
    \textbf{Round 4}  & 58.6 & 53.7 & 56.3 &  55.4& 62.1  & 57.9 & 56.9 & 79.3 & 54.1  & 82.8&  56.9 & \cellcolor{blue!15} \textbf{84.6} \\
    \textbf{Round 11} & 52.9 & 51.4 & 52.5 & 53.6 & 59.3 & 48.6 & 54.3 & 61.7 & 51.3 & 61.3& 48.6  & \cellcolor{blue!15} \textbf{80.4} \\
\specialrule{1.5pt}{\aboverulesep}{\belowrulesep}
\end{tabular}
}
\end{subtable}


\begin{subtable}[t]{\textwidth}
\caption{Comparison of scanning performance between DISTIL and alternative methods on the TrojAI benchmark for Trojaned object detection models.}
\label{table:object_detection}
\resizebox{\textwidth}{!}{%
\begin{tabular}{l c c c c c c c c c c c c}
\specialrule{1.5pt}{\aboverulesep}{\belowrulesep}
    \multirow{3}{*}{\textbf{Dataset}} & \multicolumn{12}{c}{\textbf{Method}} \\
    \cmidrule(lr){2-13}
    & \multirow{1}{*}{\textbf{NC}}
      & \multirow{1}{*}{\textbf{ABS}}
      & \multirow{1}{*}{\textbf{Pixel}}
      & \multirow{1}{*}{\textbf{\small THTP}}
      & \multirow{1}{*}{\textbf{\footnotesize SmoothInv}}
      & \multirow{1}{*}{\textbf{\small UNICORN}}
      & \multirow{1}{*}{\textbf{\footnotesize BTI-DBF}}
      & \multirow{1}{*}{\textbf{K-Arm}}
      & \multirow{1}{*}{\textbf{\small MM-BD}}
      & \multirow{1}{*}{\textbf{\small TRODO}}
      & \multirow{1}{*}{\textbf{UMD}}
      & \textbf{\small DISTIL} \\
      \specialrule{1.5pt}{\aboverulesep}{\belowrulesep}
    \textbf{TrojAI-Object Detection} & 51.1 & 46.8 & 37.0 &  54.3& 52.9 & 52.5 & 52.8 & 46.3 & 48.5 & 52.0 & 53.3 & \cellcolor{blue!15} \textbf{63.7} \\
\specialrule{1.5pt}{\aboverulesep}{\belowrulesep}
\end{tabular}
}
\end{subtable}

\label{tab:example_random_red}
\end{table*}

%% file: sec/5_Experiment.tex
\input{Tables/table_4}

\input{Tables/table_2}

\input{Tables/table_3}

\input{sec/Fig_3}

\section{Experiments}
We evaluated our method using challenging open-access Trojan scanning datasets, including BackdoorBench~\cite{wu2022backdoorbench} and TrojAI~\cite{trojai_software_framework}. We compared DISTIL against various post-training Trojan defense methods on different tasks, including such as Trojan model detection, target class identification, and Trojan model mitigation.

\textbf{Experimental Setup and Evaluation Details.}\label{sec:Experimental_Setup}  Table ~\ref{table:backdoorbench_model_detection} provides a comprehensive comparison between our method and alternative approaches for distinguishing Trojaned classifiers from clean models. Each row corresponds to a specific attack method employed to compromise the Trojaned classifier, ranging from representative to advanced attacks. The Trojaned classifiers utilized in these experiments span multiple architectures, including ResNet, VGG, ViT-B16, and ConvNeXt Tiny. We tested Trojaned models from BackdoorBench alongside 100 different clean models, since the BackdoorBench dataset originally included only one clean model per architecture.

Table~\ref{table:trojai_model_detection} details our experimental results on various rounds of the TrojAI dataset. Notably, these rounds become progressively more challenging, incorporating sophisticated label mapping techniques (such as one-to-one mappings) and diverse training strategies for poisoning Trojaned classifiers, including different adversarial training approaches.

Table~\ref{table:object_detection} illustrates DISTIL's performance on the TrojAI Round 10 dataset, which involves object detection tasks. This evaluation includes Trojaned and clean object detection models based on FastRCNN and SSD architectures. These results collectively highlight the robustness of our detection framework against a broad spectrum of backdoor attack paradigms. Figure~\ref{fig:fig3} further demonstrates the impact of trigger injection on object detection by showing how the addition of the reconstructed trigger causes significant misclassification and localization errors, underscoring the vulnerability of these models to Trojan attacks.

Table~\ref{table:mitigation} presents our results on Trojan classifiers from the BackdoorBench dataset, specifically focusing on the CIFAR-10 mitigation task. In this experiment, we fine-tuned the Trojan-infected models using the generated triggered data combined with 1\% of clean training data (a common mitigation protocol), employing cross-entropy loss. Our approach aims to address and correct the model's biased learning caused by incorrect shortcut patterns (i.e., triggers). The results demonstrate the superiority of our method in accurately reconstructing trigger patterns closely resembling the original triggers used to Trojan the classifiers. For fairness in evaluation, comparisons were restricted exclusively to RET-based methods. 

Table~\ref{table:target_class_detection} shows the performance of our method in predicting the target classes of Trojan classifiers subjected to various attacks from the BackdoorBench dataset. Specifically, we evaluate classifiers compromised by all-to-one attacks, highlighting the effectiveness and robustness of our approach under different attack scenarios.

\textbf{Evaluated Methods and Implementation Details.}
Our study focuses on RET; therefore, our comparisons are on representative and recent RET methods, including NC~\cite{wang2019neural}, ABS~\cite{ABS}, Pixel~\cite{BetterTrigger}, THTP\cite{Topo}, SmoothInv~\cite{sun2023single}, Unicorn~\cite{wang2023unicorn}, BTI-DBF~\cite{xu2024towards}, K-Arm~\cite{kARM}, and UMD~\cite{umd}. Additionally, we include MM-BD \cite{MMBD} and TRODO,recent~\cite{mirzaei2024scanning}, methods specifically designed for Trojan scanning without explicit trigger estimation. In Tables~\ref{tab:consolidated} and~\ref{table:target_class_detection}, we compare RET methods that generate informative triggers for both mitigation and target class prediction tasks. To ensure a fair comparison, we evaluated only their estimated triggers, excluding any additional components or strategies that might otherwise skew the results.

As our default backbone, we utilized the pre-trained lightweight guided diffusion model from OpenAI~\cite{nichol2021glide}, which was trained on approximately 64 million images. This model employs a fast sampling strategy that significantly improves the efficiency of the pipeline. When using the diffusion model, we set the number of sampling steps ($T$) to 50 to improve time efficiency while keeping the other hyperparameters at their default settings. We note that even when using lighter pre-trained diffusion models, such as those trained on 1 million ImageNet images, DISTIL achieves consistent performance with only a minor drop (see ablation study setups E and F in Section \ref{ablation_section} for details).

DISTIL generates triggers by aiming for high classifier confidence toward the target class. If this criterion is not initially met, we repeat the trigger generation process up to a maximum of 5 iterations, a value chosen based on empirical observations. For selecting the remaining hyperparameters, we used the models from the TrojAI training rounds.

In Appendix we present the pseudo‑code, the standard deviations of our results, additional visualizations of synthesized triggers, a background review, and DISTIL’s performance under All‑to‑All attacks. Because BackdoorBench and TrojAI focus mainly on all‑to‑one and  one‑to‑one, we also evaluate DISTIL in an all‑to‑all label mapping scenario; the results are presented in Table \ref{table:all_to_all_distill}.

\textbf{Results Analysis.}
\label{sec:Results_Analysis}DISTIL achieves an average performance of 88.5\% on BackdoorBench, 81.4\% on TrojAI, and 63.7\% on Trojaned object detection scanning task. Notably, DISTIL surpasses alternative methods by high margins, achieving up to \textbf{7.1\% } higher accuracy on the BackdoorBench dataset and a \textbf{9.4\%} improvement on Trojaned object detection model scanning. These results underscore  effectiveness as a robust scanning approach that is applicable in diverse tasks. Furthermore, DISTIL significantly lowers the Attack Success Rate (ASR) to just 8.6\%, while concurrently improving the target class prediction accuracy to 72.0\% on the GTSRB dataset.
While DISTIL performance gap is smaller on TrojAI rounds, which mainly involve Trojaned classifiers with patch-shaped triggers similar to Badnet \cite{gu2019badnets} attacks, DISTIL consistently achieves high performance on the broader BackdoorBench dataset. This highlights its ability to detect diverse Trojan attacks.

%% file: Tables/table_4.tex
\begin{table*}[ht]
\centering
\caption{Ablation study showing model accuracy (\%) when each component is individually excluded or replaced, with all other components held constant.}
\resizebox{\textwidth}{!}{
\begin{tabular}{@{} c c c c c c c c cc c c c c c} 
\specialrule{1.5pt}{\aboverulesep}{\belowrulesep}
    
\multirow{3}{*}{\textbf{Setup}} 
& \multicolumn{6}{c}{\textbf{Components}} 
& 
& \multicolumn{6}{c}{\textbf{Dataset}}\\

\cmidrule(lr){2-8} \cmidrule(lr){10-15}

&{\footnotesize  Data Supervision}
& Noise 
& {\footnotesize Fast Class} 
& {\footnotesize Training\&Hyper }
& {\footnotesize Dif. Model-1}
& {\footnotesize Dif. Model-2}
& {\footnotesize Dif. Model-3}
&
& \multirow{2}{*}{\textbf{\footnotesize  Round0}} 
& \multirow{2}{*}{\textbf{\footnotesize Round1}} 
& \multirow{2}{*}{\textbf{\footnotesize Round2}} 
& \multirow{2}{*}{\textbf{\footnotesize Round3}} 
& \multirow{2}{*}{\textbf{\footnotesize Round4}}
& \multirow{2}{*}{\textbf{\footnotesize Round11}}\\

& {\footnotesize for RET}
& {\footnotesize Injection}
&  {\footnotesize Pairing}
& {\footnotesize Selection}
& \cite{nichol2021glide}
& \cite{dhariwal2021diffusion}
& \cite{song2020score}
&
& 
& 
& 
& 
& 
& \\

\specialrule{1.5pt}{\aboverulesep}{\belowrulesep}
\vspace{1mm}

\textbf{A} 
& \checkmark & \checkmark & - & \checkmark & - & - & - 
& \vline
& 74.5 & 73.0 & 65.2 & 64.9 & 60.5 & 57.4\\
\specialrule{0.05pt}{\aboverulesep}{\belowrulesep}
\vspace{1mm}

\textbf{B} 
& - & - & - & \checkmark & \checkmark  & - & -
& \vline
& 81.9 & 80.5 & 76.3 & 74.4 & 81.6 & 76.9 \\
\specialrule{0.05pt}{\aboverulesep}{\belowrulesep}
\vspace{1mm}
\textbf{C} 
&- & \checkmark & - & - & \checkmark & - & -
& \vline 
& 80.6 & 76.4 & 72.8 & 74.1 & 78.0 & 73.3 \\
\specialrule{0.05pt}{\aboverulesep}{\belowrulesep}
\vspace{1mm}
\textbf{D} 
& - & \checkmark & \checkmark & \checkmark & \checkmark & - & -
& \vline 
& 78.0 & 79.1 & 73.9 & 75.9 & 82.3 & 75.6 \\
\specialrule{0.05pt}{\aboverulesep}{\belowrulesep}
\vspace{1mm}

\textbf{E}  
& - & \checkmark & - & \checkmark & - &  \checkmark &-
& \vline 
& 81.9 & 80.3 & 78.4 & 78.2 & 81.0 & 78.4 \\
\specialrule{0.05pt}{\aboverulesep}{\belowrulesep}
\vspace{1mm}
\textbf{F}\textit{\tiny  }  
& - & \checkmark & - & \checkmark & - & - & \checkmark
& \vline 
& 78.6 & 74.2 & 73.8 & 75.1 & 80.3 & 77.0 \\
\specialrule{1.5pt}{\aboverulesep}{\belowrulesep}
\vspace{1mm}

\textbf{G}\textit{\tiny (Ours)}  
& - & \checkmark & - & \checkmark & \checkmark & - & -
& \vline 
& 83.1 & 82.9 & 79.5 & 78.4 & 84.6 & 80.4 \vspace{1mm} \\
\specialrule{0.05pt}{\aboverulesep}{\belowrulesep}
\vspace{1mm}
\textbf{H}\textit{\tiny (Ours+Data)} 
& \checkmark & \checkmark & - & \checkmark & \checkmark & - & -
& \vline 
& 84.5 & 83.6 & 82.4 & 81.8 & 86.0 & 83.9 \\

\specialrule{1.5pt}{\aboverulesep}{\belowrulesep}
\end{tabular}
}
\label{table:ablation_study}
\end{table*}

%% file: Tables/table_2.tex
\begin{table*}[ht!]
\centering
\caption{Mitigation results on CIFAR-10 Trojaned classifiers from BackdoorBench. We report post-fine-tuning classification accuracy (ACC $\uparrow$) and attack success rate (ASR $\downarrow$) across various Trojan attack scenarios, compared to the \emph{original} (unmitigated) models.}
\label{table:mitigation}
\resizebox{\textwidth}{!}{%
\begin{tabular}{ll cc cc cc cc cc cc cc cc cc}
\specialrule{1.5pt}{\aboverulesep}{\belowrulesep}
    \multirow{3}{*}{\textbf{Dataset}} & 
    \multirow{3}{*}{\textbf{Attack}} & 
    \multicolumn{16}{c}{\textbf{Method}} \\
    \cmidrule(lr){3-18}
    & &
    \multicolumn{2}{c}{\textbf{\textit{Original}}} &
    \multicolumn{2}{c}{\textbf{NC}} &
    \multicolumn{2}{c}{\textbf{Pixel}} &
    \multicolumn{2}{c}{\textbf{THTP}} &
    \multicolumn{2}{c}{\textbf{\small SmoothInv}} &
    \multicolumn{2}{c}{\textbf{Unicorn}} &
    \multicolumn{2}{c}{\textbf{\small BTI-DBF}} &
    \multicolumn{2}{c}{\begin{tabular}{c}
        \textbf{\small DISTIL}\\
        \textbf{\textit{\small (Ours)}}
    \end{tabular}} \\
    \cmidrule(lr){3-4}\cmidrule(lr){5-6}\cmidrule(lr){7-8}\cmidrule(lr){9-10}%
    \cmidrule(lr){11-12}\cmidrule(lr){13-14}\cmidrule(lr){15-16}\cmidrule(lr){17-18}
    & & 
\textbf{ACC.}$\uparrow$ & \textbf{ASR}$\downarrow$ &
\textbf{ACC.}$\uparrow$ & \textbf{ASR}$\downarrow$ &
\textbf{ACC.}$\uparrow$ & \textbf{ASR}$\downarrow$ &
\textbf{ACC.}$\uparrow$ & \textbf{ASR}$\downarrow$ &
\textbf{ACC.}$\uparrow$ & \textbf{ASR}$\downarrow$ &
\textbf{ACC.}$\uparrow$ & \textbf{ASR}$\downarrow$ &
\textbf{ACC.}$\uparrow$ & \textbf{ASR}$\downarrow$ &
\textbf{ACC.}$\uparrow$ & \textbf{ASR}$\downarrow$ \\
\specialrule{1.5pt}{\aboverulesep}{\belowrulesep}

 \multirow{10}{*}{ \rotatebox[origin=c]{90}{\textbf{CIFAR-10}}}
 
    & \textbf{BadNets}  
        & 91.7 & 94.4
        & 86.3 & 9.5
        & 88.0 & 15.2
        & 87.2 & 10.9
        & 86.2 & \cellcolor{blue!15}\textbf{7.4}
        & 89.0 & 12.2
        & 91.1 & 8.7
        & 90.3 & 8.6 \\
    & \textbf{Blended}
        & 93.6 & 99.7
        & 85.9  & 8.9
        & 89.6 & 12.4
        & 87.8 & 7.8
        & 92.5& 6.8
        & 90.6 & 10.4
        & 90.8 & 6.5
        & 89.1 & \cellcolor{blue!15}5.3 \\

    & \textbf{BPP}
        & 93.8 & 99.8
        & 87.9  & 97.6
        & 91.3 & 82.0
        & 89.6 & 89.8
        & 88.6 & 90.8
        & 92.3 & 81.0
        & 89.5 & 12.4
        & 88.4 & \cellcolor{blue!15} \textbf{9.0} \\
    & \textbf{Inputaware}
        & 89.6 & 94.6
        & 85.0 & 38.4
        & 86.7 & 22.1
        & 85.9 & 30.3
        & 84.0 & 31.3
        & 87.7 & 25.6
        & 86.9 & 10.8
        & 87.2 & \cellcolor{blue!15}\textbf{7.4}\\
    & \textbf{LC}
        & 84.5 & 99.9
        & 79.3 & 18.7
        & 78.4 & 16.2
        & 78.9 & 17.5
        & 80.9 & 18.5
        & 79.4 & 15.6
        & 82.0 & 12.3
        & 86.5 & \cellcolor{blue!15} \textbf{10.7} \\
    & \textbf{LF}
        & 89.4 & 30.2
        & 83.4 & 9.1
        & 86.2 & 13.9
        & 84.8 & 6.5
        & 83.8 & 7.5
        & 87.2 & 12.9
        & 85.7 & 8.1
        & 91.6 & \cellcolor{blue!15}\textbf{5.6} \\

    & \textbf{SIG}
        & 84.5 & 97.1
        & 78.2 & 32.8
        & 80.7 & 14.3
        & 79.5 & 19.6
        & 78.5 & 20.6
        & 81.7 & 20.3
        & 80.2 & 14.8
        & 82.5 & \cellcolor{blue!15}\textbf{12.8} \\
    & \textbf{SSBA}
        & 92.8 & 97.1
        & 89.1 & 14.2
        & 91.4 & 9.0
        & 90.3 & 15.6
        & 89.3 & 8.1
        & 91.4 & 15.0
        & 88.4 & 12.6
        & 86.1 & \cellcolor{blue!15}\textbf{7.9} \\
    & \textbf{TrojanNN}
        & 93.4 & 100.0
        & 90.7 & 11.6
        & 88.5 & 10.6
        & 89.6 & 13.4
        & 90.6 & 12.1
        & 89.5 & 15.6
        & 87.6 & 9.4
        & 91.9 & \cellcolor{blue!15}\textbf{8.2} \\
    & \textbf{WaNet}
        & 87.8 & 85.7
        & 91.5 & 14.3
        & 86.0 & 15.7
        & 88.8 & 19.6
        & 87.8 & 16.4
        & 85.0 & 14.7
        & 85.3 & \cellcolor{blue!15}\textbf{9.2}
        & 86.1 & 10.5
         \\

\specialrule{1.5pt}{\aboverulesep}{\belowrulesep}
\end{tabular}
}
\label{tab:consolidated}
\end{table*}

%% file: Tables/table_3.tex
\begin{table}[ht]
\centering
\caption{Target‐class prediction accuracy of RET methods on BackdoorBench models Trojaned with various backdoor attacks on the GTSRB dataset. Each row corresponds to a different attack type, and each column shows the accuracy achieved by existing RET baselines versus our proposed approach.  }
\resizebox{\columnwidth}{!}{%
\begin{tabular}{llccccccc}
\specialrule{1.5pt}{\aboverulesep}{\belowrulesep}
\multirow{3}{*}{\textbf{ Dataset}} & \multirow{3}{*}{\textbf{Attack}} & \multicolumn{7}{c}{\textbf{Method}} \\
\cmidrule(lr){3-9}
& & \multirow{2}{*}{\textbf{NC}} & \multirow{2}{*}{\textbf{Pixel}} & \multirow{2}{*}{\textbf{THTP}} & \multirow{2}{*}{\textbf{\footnotesize SmoothInv}} & \multirow{2}{*}{\textbf{Unicorn}} & \multirow{2}{*}{\textbf{\small BTI-DBF}} & \textbf{\small DISTIL} \\
& & & & & & & & \textbf{\textit{\small (Ours)}} \\
\specialrule{1.5pt}{\aboverulesep}{\belowrulesep}
 \multirow{11}{*}{ \rotatebox[origin=c]{90}{\textbf{GTSRB}}}&

\textbf{BadNets} & 0.65 & 0.65 & 0.60 & \cellcolor{blue!15}\textbf{0.80} & 0.75 & \cellcolor{blue!15}\textbf{0.80} & \cellcolor{blue!15}\textbf{ 0.80}\\
& \textbf{Blended} & 0.65 & 0.55 & 0.45 & 0.55 & 0.65 & 0.70& \cellcolor{blue!15}\textbf{0.85}\\
& \textbf{BPP} & 0.45 & 0.50& 0.55 & 0.45 & 0.40 & 0.50 & \cellcolor{blue!15}\textbf{0.65}\\
& \textbf{\footnotesize Inputaware} & 0.30 & 0.25 & 0.20 & 0.35& 0.45 & 0.65 & \cellcolor{blue!15}\textbf{0.70}\\
& \textbf{LC} & 0.45 & 0.40 & 0.45 & 0.60 & 0.65 & 0.60 & \cellcolor{blue!15}\textbf{0.70}\\
& \textbf{LF} & 0.40 & 0.45 & 0.55 & 0.55 & 0.30 & \cellcolor{blue!15}\textbf{0.65} & \cellcolor{blue!15}\textbf{0.65}\\
& \textbf{LIRA} & 0.55 & 0.50 & 0.45 & 0.40 & 0.55 & 0.60 & \cellcolor{blue!15}\textbf{0.75}\\
& \textbf{SIG} & 0.20 & 0.25 & 0.30 & 0.20 & 0.35 & 0.65& \cellcolor{blue!15}\textbf{0.70}\\
& \textbf{SSBA} & 0.35 & 0.30 & 0.25 & 0.45 & 0.55 & 0.60 & \cellcolor{blue!15}\textbf{0.65}\\
& \textbf{TrojanNN} & 0.40 & 0.45 & 0.50 & 0.55 & 0.50 & 0.75 & \cellcolor{blue!15}\textbf{0.80}\\
& \textbf{WaNet} & 0.25 & 0.35 & 0.20 & 0.30& 0.45 & 0.65 & \cellcolor{blue!15}\textbf{0.75}\\
\specialrule{1.5pt}{\aboverulesep}{\belowrulesep}
\end{tabular}
}

\label{table:target_class_detection}
\end{table}

%% file: sec/Fig_3.tex
\begin{figure*}[t]
  \centering
  \includegraphics[width=\linewidth]{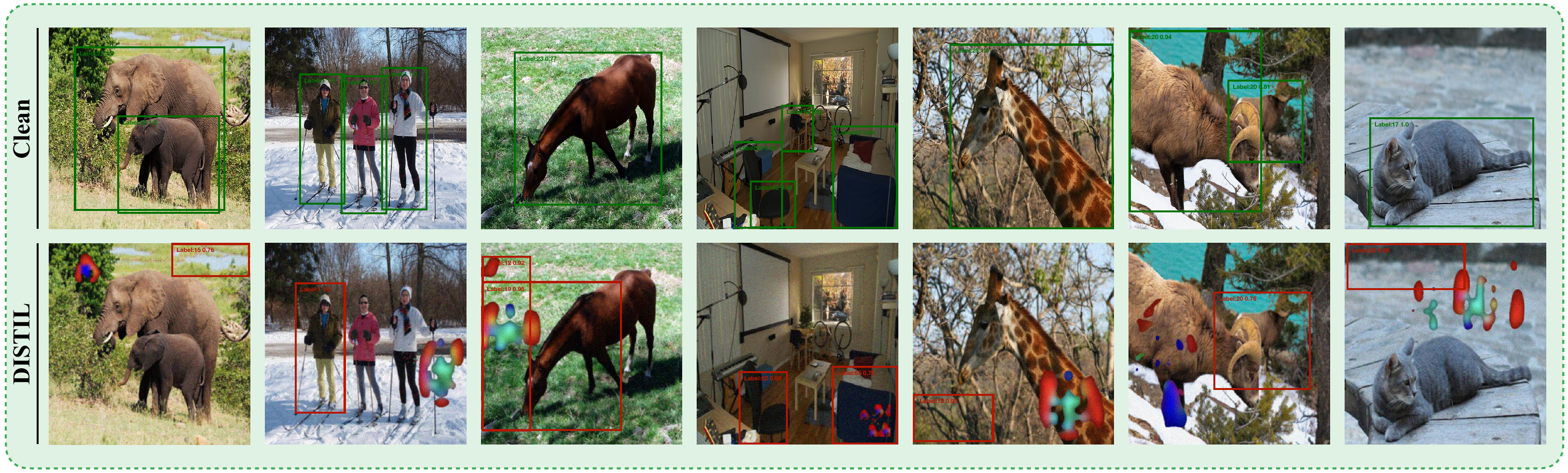}
  \caption{\textbf{DISTIL on object‑detection models.}
Each pair shows (top row) a clean input with correct detections (green boxes) and the same input (bottom row) after injecting the trigger recovered by DISTIL (red boxes). The reconstructed trigger not only flips the model’s prediction from the true class to the attacker’s target class but also drags the bounding-box center toward a pre-selected corner, producing simultaneous misclassification and mislocalization. This visualization demonstrates DISTIL’s ability to generalize from image classification to object detection task. }
  \label{fig:fig3}
\end{figure*}

%% file: sec/6_Ablation_Study.tex
\section{Ablation Study} \label{ablation_section}

We performed an ablation study across multiple TrojAI rounds (Table~\ref{table:ablation_study}) to isolate the contributions of each component in DISTIL. Our default configuration, Setup G, employs no data supervision for RET, injects uniform noise into the classifier input, selects hyperparameters based on training on a subset of the TrojAI training data, and uses the GLIDE \cite{nichol2021glide} as the generative backbone. In Setup H, we augment the default setting by introducing clean training data for RET, enabling the hybrid conditioning described in Equation \ref{Equation:hybrid}. In Setup A, we remove the diffusion model altogether and optimize the objective directly in pixel space, thereby demonstrating the significance of diffusion modeling in synthesizing discriminative triggers. In Setup B, we discard the noise-injection strategy while keeping every other component unchanged, illustrating how noise injection helps avert adversarial artifacts. In  Setup C, we examine the role of hyperparameter tuning by abandoning training-based hyperparameter selection and instead using fixed values of $\lambda_{1}=0.3$ (noise injection strength) and $\lambda_{2}=0.95$ (classification confidence threshold); these values were chosen to avoid excessive distortion of important input features and to ensure that the classifier has high confidence in attributing the crafted trigger to the target class, a setting that underscores DISTIL's robustness to suboptimal hyperparameters. In Setup D, we adopt our fast RET strategy for trigger generation, which reduces the computational complexity from $\mathcal{O}(K^2)$ to $\mathcal{O}(K)$, illustrating DISTIL's capacity to balance efficiency with strong detection performance. In Setup E, we replace the default diffusion model with an ImageNet-pretrained one proposed by \cite{dhariwal2021diffusion}, and in Setup F, we use a score-based diffusion model pretrained on LSUN \cite{song2020score}. Both variations reveal only modest changes in performance, confirming that while the backbone diffusion model and its pretraining data matter, DISTIL retains a high level of effectiveness. 

%% file: sec/7_Conclusion.tex
\section{Conclusions}
We introduced DISTIL, a novel diffusion-based method for accurately reconstructing interpretable Trojan triggers without needing training data. By integrating classifier-guided diffusion with injected noise, DISTIL effectively distinguishes genuine triggers from adversarial noise, significantly reducing false positives in detection. Extensive evaluations across multiple architectures and benchmarks confirmed DISTIL's robust performance in detecting, predicting, and mitigating Trojan attacks, highlighting its practical utility for enhancing model security in critical vision tasks.

%% file: sec/8_Appendix.tex
\maketitlesupplementary
\label{sec:appendix_section}

\section{Algorithm Block} \label{appendix:algorithm_block}
\input{sec/Appendix/algorithm.tex}

\section{Additional Technical Background} \label{appendix:ddpm_background}

\textbf{Denoising Diffusion Probabilistic Models (DDPMs).}  
DDPMs have emerged as a promising approach for generating high-quality data across various domains, particularly in image and video synthesis. They operate by reversing a forward process that gradually adds Gaussian noise to the data over \(T\) steps. Formally, the forward process is:
\begin{equation}
    q(x_t \mid x_{t-1}) = \mathcal{N}\bigl(x_t; \sqrt{1-\beta_t}\,x_{t-1}, \beta_t \mathbf{I}\bigr),
\label{eq:forward_process}
\end{equation}  
where \(\{\beta_t\}_{t=1}^T\) controls the noise schedule. The corresponding reverse process is learned to iteratively denoise:  
\begin{equation}
    p_{\theta}(x_{t-1} \mid x_t) = \mathcal{N}\bigl(\mu_{\theta}(x_t, t), \Sigma_{\theta}(x_t, t)\bigr).
\label{eq:reverse_process}
\end{equation}  

A neural network \(\epsilon_{\theta}\) is then trained to predict the added noise using the following objective:  
\begin{equation}
    \mathcal{L}_{\text{simple}} = \mathbb{E}_{t,\,x_0,\,\epsilon} \Bigl[\|\epsilon - \epsilon_{\theta}(x_t, t)\|^2\Bigr].
\label{eq:loss_function}
\end{equation}

Subsequent research has introduced conditioning and guidance mechanisms, enabling the model to produce outputs with specific attributes (e.g., guided by text prompts).

\input{sec/Fig_4}

\input{sec/Fig_5}

\section{Visualization of the Generated Backdoor Trigger} \label{appendix:visualization}

Figure~\ref{fig:fig4} presents the trigger patterns generated by DISTIL when applied to clean models across multiple rounds (Rounds 1, 2, 3, 4, and 11). As expected, these patterns appear noisy and lack coherent structure, demonstrating that DISTIL does not erroneously extract trigger-like features from benign systems. In contrast, Figure~\ref{fig:fig5} illustrates the trigger pattern reconstructed by DISTIL on the CIFAR-10 dataset under the BadNets attack scenario. In this setting, our latent diffusion-based approach consistently recovers clear and interpretable trigger patterns that capture the distinct characteristics of the BadNets attack. Together, these visualizations underscore DISTIL's robustness and its discriminative power in distinguishing Trojaned models from clean ones.

\section{Additional Experimental Results} \label{appendix:experimental_results}

See Tables~\ref{table:X}, and ~\ref{table:all_to_all_distill}.

\begin{table*}[ht]
\centering
\caption{Evaluation of our model's performance across various architectures using the BackdoorBench dataset.}
\label{table:X}
\resizebox{.6\textwidth}{!}{
\begin{tabular}{l c c c c}
\specialrule{1.5pt}{\aboverulesep}{\belowrulesep}
& \multicolumn{4}{c}{\textbf{Architecture}} \\
\cmidrule(lr){2-5}
\textbf{Method} & \textbf{Pre-act Resnet18} & \textbf{VGG-19 BN} & \textbf{VIT B 16} & \textbf{ConvNeXt Tiny} \\
\specialrule{1.5pt}{\aboverulesep}{\belowrulesep}
\textbf{DISTIL} & 88.5 & 89.3 & 92.8 & 87.6 \\
\specialrule{1.5pt}{\aboverulesep}{\belowrulesep}
\end{tabular}
}
\end{table*}

\begin{table*}[t]
\centering
\caption{Performance evaluation of our method across diverse attack scenarios in an all-to-all setting.}
\label{table:all_to_all_distill}
\resizebox{0.7\textwidth}{!}{
\begin{tabular}{l c c c c c c c c c c c}
\specialrule{1.5pt}{\aboverulesep}{\belowrulesep}
& \multicolumn{11}{c}{\textbf{Attack Methods}} \\
\cmidrule(lr){2-12}
\textbf{Method} & \textbf{BadNets} & \textbf{Blended} & \textbf{BPP} & \textbf{InputAware} & \textbf{LC} & \textbf{LF} & \textbf{LIRA} & \textbf{SIG} & \textbf{SSBA} & \textbf{TrojanNN} & \textbf{WaNet} \\
\specialrule{1.5pt}{\aboverulesep}{\belowrulesep}
\textbf{DISTIL} & 90.8 & 89.3 & 84.4 & 89.6 & 86.5 & 87.3 & 86.9 & 89.0 & 84.6 & 81.3 & 82 \\
\specialrule{1.5pt}{\aboverulesep}{\belowrulesep}
\end{tabular}
}
\end{table*}

\section{Details of Evaluation and Experimental Setup Implementation} 
\input{Tables/table_5}
\label{appendix:evaluation_setup}

\section*{Details for the Backdoor Attacks}

\noindent This section offers comprehensive explanations of the backdoor attacks utilized in our research.

\textbf{BadNet} \cite{gu2019badnets} introduces a hidden pattern into datasets during training, often a compact and noticeable visual element. This marker is crafted to blend into the background, avoiding suspicion while still training the AI to incorrectly label any data containing this subtle cue.

\textbf{Blended} \cite{blended} merges a faint, almost invisible signal into training visuals, keeping its presence undetectable to casual observation or basic scanning tools. By gradually linking these barely noticeable alterations to wrong predictions, the model learns to produce errors when triggered.

\textbf{SIG} \cite{sig} modifies training images by adding wave-like distortions without adjusting their categorizations. This covert strategy bypasses traditional defenses that monitor mismatched labels, allowing the hidden flaw to persist unnoticed.

\textbf{BPP} \cite{bpp} employs data compression techniques and adversarial training to insert triggers directly into the numeric values of individual pixels. These microscopic changes are nearly impossible to visually identify or computationally trace, hiding attacks within the image's core structure.

\textbf{Input-aware} attacks \cite{inputaware} create adaptive triggers that morph depending on the unique characteristics of each data sample. The backdoor remains dormant until specific input criteria—predefined by the attacker—are met, enhancing its ability to evade discovery.

\textbf{WaNet} \cite{nguyen2021wanet} applies barely detectable geometric distortions to images, bending their spatial features in subtle ways. These warped elements serve as invisible keys that bypass human vision while reliably tricking the model.

\textbf{SSBA} \cite{ssba} generates unique, undetectable markers for every training sample by weaving triggers into the image’s inherent textures and patterns. This individualized approach complicates large-scale detection efforts that rely on universal trigger signatures.

\textbf{Color} \cite{color} manipulates hue and saturation channels in images to create chromatic triggers. These alterations fly under the radar of standard visual audits but condition models to recognize and act on specific color shifts.

\textbf{Label Consistency Attack (LC)}\cite{turner2019labelconsistency} embeds triggers into training samples without disrupting their apparent class labels, ensuring poisoned data remains label-consistent. This is achieved by designing triggers that minimally alter features relevant to the true class while conditioning the model to associate the trigger with the attacker’s desired output.

\textbf{TrojanNN} \cite{Trojannn} preprocesses triggers to maximize activation of specific neurons critical to the model’s decision-making. By embedding these optimized triggers, the attack ensures misclassification of triggered inputs while maintaining high accuracy on clean data. The triggers exploit the neural network’s internal structure, making them difficult to detect or reverse-engineer through standard methods.

\textbf{Label Flipping (LF)} \cite{rosenfeld2020certified} manipulates the labels of training samples containing a specific trigger, causing the model to associate the trigger with an incorrect class. This attack ensures normal behavior on clean inputs while misclassifying triggered inputs, maintaining stealth by using inconspicuous triggers (e.g., clapping sounds in audio systems) that blend naturally with the data.

\subsection{Metrics}

We primarily use ACC to present our results, aligning with its widespread adoption for method comparison in the existing literature. Detailed explanations of all the metrics used are provided below.

\noindent\textbf{ACC.}\; Accuracy is a fundamental metric used to evaluate the performance of classification models, representing the proportion of correctly predicted instances out of the total number of instances. It is calculated as the ratio of true positives (correctly predicted positive instances) and true negatives (correctly predicted negative instances) to the sum of all predictions, including false positives and false negatives.

\subsection{Time Complexity}

Table \ref{table:trojai_time_complexity} presents the time complexity of DISTIL on the TrojAI dataset. Experiments were performed using an RTX A5000 GPU.

\begin{table*}[t]
\centering
\caption{Average time complexity (hours) of DISTIL-Fast and DISTIL on TrojAI models.}
\label{table:trojai_time_complexity}
\resizebox{0.6\textwidth}{!}{%
\begin{tabular}{l c c c c c c}
\specialrule{1.5pt}{\aboverulesep}{\belowrulesep}
    \multirow{2}{*}{\textbf{Method}} & \multicolumn{6}{c}{\textbf{Dataset}} \\
\cmidrule(lr){2-7}
    & \textbf{Round 0} & \textbf{Round 1} & \textbf{Round 2} & \textbf{Round 3} & \textbf{Round 4} & \textbf{Round 11} \\
\specialrule{1.5pt}{\aboverulesep}{\belowrulesep}
\textbf{DISTIL-Fast}  & 0.01 & 0.01 & 0.04 & 0.03 & 0.07 & 0.2 \\
\textbf{DISTIL}  & 0.06 & 0.06 & 0.5 & 0.6 & 2.1 & 14.4 \\
\specialrule{1.5pt}{\aboverulesep}{\belowrulesep}
\end{tabular}
}
\end{table*}

\section{Previous Trojan Scanning methods}
\label{appendix:other_works}

\subsection*{Review of the Methods}

Trigger estimation plays several roles in defense strategies against Trojan attacks. The most important is determining whether a model is benign or Trojaned, a task that closely resembles out-of-distribution detection \cite{salehi2021unified,mirzaei2022fake,mirzaei2025contrastiveteacherstudentframeworknovelty,mirzaei2024universal,mirzaei2024adversarially,mirzaeirodeo,mirzaei2024killing,sträter2024generaladanomalydetectiondomains,mirzaei2025mitigatingspuriousnegativepairs,salehi2025cranecontextguidedpromptlearning,nafez2025patchguardadversariallyrobustanomaly}. In the following, we provide a brief review of related work.

\textbf{NC}\quad NC Neural Cleanse \cite{wang2019neural} is a technique designed to scan models by reverse engineering potential triggers and detecting outlier perturbations. NC suffers from significant computational overhead, is highly sensitive to trigger complexity, and may result in false positives. Additionally, its design is optimized primarily for handling all-in-one attack scenarios. 

\textbf{Pixel}\quad Better Trigger Inversion Optimization \cite{BetterTrigger} in Backdoor Scanning introduces an improved approach for reverse-engineering backdoor triggers by refining the optimization process. Unlike previous methods that struggle with fragmented or overly complex perturbation patterns, this technique optimizes trigger inversion by leveraging structured constraints to enhance efficiency and accuracy. By incorporating a refined objective function and optimization strategy, it reduces computational overhead while achieving higher fidelity in reconstructing backdoor triggers. However, the method may still face challenges in handling highly adaptive or distributed trigger patterns and could require careful parameter tuning for different attack scenarios.

\textbf{BTI-DBF}\quad BTI-DBF \cite{xu2024towards} presents a backdoor defense strategy by decoupling benign features to isolate backdoor triggers, contrasting traditional methods that directly approximate backdoor patterns. The approach involves two steps: optimizing the model to rely exclusively on benign features for accurate predictions on clean samples while rendering residual (backdoor) features ineffective and training a generator to align benign features between clean and poisoned samples while amplifying differences in backdoor-related features. Leveraging this decoupling, the proposed BTI module facilitates backdoor removal through fine-tuning with relabeled poisoned data and input purification by approximating the inverse of the backdoor generation process. Both defenses are iteratively refined by updating their generators based on the performance of their respective outputs.

\textbf{TRODO}\quad TRODO \cite{mirzaei2024scanning} leverages the concept of ``blind spots," which are regions where Trojaned classifiers erroneously identify out-of-distribution (OOD) samples as in-distribution (ID). The methodology involves adversarially shifting OOD samples toward the in-distribution and observing the model's classification behavior. An increased likelihood of perturbed OOD samples being classified as ID serves as a signature for Trojan detection. This approach does not require knowledge of the specific Trojan attack method or the label mapping, making it both Trojan and label mapping agnostic. However, this method is not without its limitations. Its effectiveness depends heavily on the quality and diversity of out-of-distribution (OOD) samples; poor or unrepresentative samples may result in undetected Trojans. Furthermore, the success of Trojan detection relies on the chosen perturbation technique, and weak or ill-suited methods could fail to expose Trojan behavior.

\textbf{SmoothInv}\quad The paper introduces SmoothInv \cite{sun2023smoothinv}, a backdoor inversion method designed to recover backdoor patterns in neural networks using only a single clean image. The method transforms a backdoored classifier into a robust smoothed classifier by applying random Gaussian perturbations to the input image, optionally followed by diffusion-based denoising. Using projected gradient descent, SmoothInv synthesizes a perturbation guided by gradients from this robust classifier to reconstruct the hidden backdoor pattern effectively. Unlike traditional inversion methods that require numerous images and complex regularizations, SmoothInv simplifies optimization and achieves high accuracy and visual fidelity to the original backdoor patterns. However, the main limitation of this approach is its ineffectiveness against advanced or subtle backdoors beyond simple patch-based attacks, such as image warping, adaptive imperceptible perturbations, or Instagram filter-based triggers.

\textbf{FeatureRE}\quad Rethinking the Reverse-engineering of Trojan Triggers \cite{wang2022rethinking} is a method that reexamines traditional trigger reverse-engineering by shifting the focus from static input-space constraints to the exploitation of feature space properties. This approach leverages the insight that both input-space and feature-space Trojans manifest as hyperplanes in the model’s feature space. By incorporating feature space constraints into the reverse-engineering process, the method can reconstruct dynamic, input-dependent triggers. This method comes with several limitations. First, similar to many reverse-engineering approaches, it requires access to a small set of clean samples, which may not always be available in practical scenarios. Second, the optimization process involves multiple hyperparameters that require careful tuning.

\textbf{THTP}\quad Trigger Hunting with a Topological Prior \cite{hu2022trigger} for Trojan Detection is a method that integrates topological data analysis into the reverse-engineering of Trojan triggers. The approach leverages global topological features to capture and differentiate anomalous triggers. By enforcing a topological prior, the method enhances the detection of both static and dynamic (input-dependent) triggers. However, the integration of topological computations introduces additional computational overhead and increases sensitivity to hyperparameter settings. Moreover, while the approach improves robustness against irregular trigger patterns, it may face scalability challenges in models with complex or multiple overlapping Trojan patterns.

\textbf{UNICORN}\quad Trigger Inversion Framework \cite{wang2023unicorn} proposes a generalized backdoor detection approach by formalizing triggers as perturbations in transformable input spaces via an invertible function. Unlike existing methods, it jointly optimizes trigger parameters (mask, pattern) and invertible function to identify optimal injection spaces, exploiting disentangled compromised and benign activation vectors. However, the framework's computational overhead from multi-space optimization and reliance on activation disentanglement assumptions may limit scalability and robustness against advanced multi-trigger attacks.

\textbf{MM-BD}\quad Maximum Margin Backdoor Detection \cite{MMBD} is a technique aimed at identifying backdoor attacks in neural networks, irrespective of the type of backdoor pattern used. The approach works by calculating a maximum margin statistic for each class using gradient ascent from various random starting points, all without requiring clean data samples. These statistics are then applied within an unsupervised anomaly detection system to pinpoint backdoor attacks. However, the method has notable limitations: it tends to produce a high rate of false positives when dealing with datasets containing few classes, and it has difficulty detecting attacks involving multiple target labels, where different source classes may correspond to distinct target classes. Furthermore, MM-BD's performance is considerably weakened when faced with an adaptive attacker who alters the learning process.

\textbf{ABS}\quad Artificial Brain Stimulation \cite{ABS} identifies backdoors in neural networks by activating specific neurons and measuring how their stimulation alters model predictions. Suspect neurons that disproportionately influence a target class are flagged, and corresponding input patterns are reconstructed to verify malicious activity. While effective in controlled settings, the approach demands extensive computational resources and relies on assumptions about trigger characteristics. Its efficacy diminishes with sophisticated attacks, particularly those beyond single-target scenarios. False alarms may arise when legitimate neurons exhibit strong class correlations, reducing reliability in diverse applications.

\textbf{TABOR}\quad TABOR \cite{TABOR} detects Trojans in DNNs by treating trigger identification as a constrained optimization task. It employs a loss function augmented with interpretability-driven penalties to narrow down plausible trigger candidates. Although this structured approach enhances precision for geometric or symbolic triggers, computational costs remain prohibitive. Irregular or non-traditional trigger patterns may evade detection, as the method’s design prioritizes simplicity over adaptability.

\textbf{PT-RED}\quad PT-RED \cite{PTRED} uncovers backdoors post-training by generating subtle input alterations that force misclassifications. The technique prioritizes small, localized trigger patterns linked to a single target class. However, the iterative optimization required to synthesize these perturbations is resource-intensive. Its narrow focus on basic attack types limits applicability to complex, multi-class scenarios, and scalability becomes problematic with larger models.

\textbf{K-ARM}\quad K-ARM \cite{kARM} adopts a trial-and-error strategy, dynamically prioritizing class labels for trigger exploration using principles from resource allocation theory. This adaptive sampling reduces redundant computations compared to brute-force methods. Nevertheless, processing overhead persists, and the approach falters when attacks involve multiple targets or intricate trigger designs, restricting its utility to simpler threat models.

\textbf{UMD}\quad Unsupervised Model Detection \cite{umd} targets multi-class backdoor attacks by crafting input perturbations for every possible class pairing. It quantifies cross-pattern consistency (e.g., whether a trigger for one class pair affects others) and applies statistical anomaly detection to identify compromises. While effective for single-trigger threats, the method’s reliance on exhaustive trigger synthesis and pairwise analysis strains computational resources. Scalability wanes with large-scale models or datasets, and its accuracy degrades if multiple distinct triggers coexist.

\section{Details about the Benchmarks and Datasets} \label{appendix:datasets}

We provide a brief explanation of the datasets we used.

\textbf{BackdoorBench Benchmark.} BackdoorBench \cite{wu2022backdoorbench} is a comprehensive benchmark for backdoor learning, providing a standardized platform for evaluating backdoor attacks and defenses. It consists of four modules: input, attack, defense, and evaluation and analysis. The attack module offers sub-modules for implementing data poisoning and training controllable attacks, while the defense module provides sub-modules for implementing backdoor defenses. BackdoorBench has been widely used in evaluating various backdoor defense methods and has been used as a reference in several papers.

\textbf{TrojAI Benchmark.} TrojAI Benchmark \cite{trojai_software_framework} is a dedicated dataset curated for evaluating the resilience of neural network models against Trojan attacks. It comprises a diverse collection of pre-trained models, including both clean and Trojaned networks. The benchmark spans various network architectures and incorporates multiple Trojan injection strategies, thereby simulating a wide range of real-world attack scenarios. Below is an overview of the details for each round of the TrojAI benchmark for the classification task. 

\textbf{Round 0 (Dry Run).} This initial round is designed as a preliminary test. It consists of 200 CNN models trained on human-level image classification on synthetic traffic sign data (with 5 classes), where half of the models are poisoned with an embedded trigger, which causes misclassification of the images when the trigger is present. 

\textbf{Round 1.} This round comprised 1,000 CNN models, specifically Inception-v3, DenseNet-121 and ResNet50 architectures, trained to classify synthetic street signs superimposed on road backgrounds with 50\% containing hidden Trojans triggered by polygon-based patterns of uniform color and varying shapes/sizes (2- 25\% of the surface area of the target object). This round covers all-to-one types of triggers. The test dataset consists of 100 models. 

\textbf{Round 2} Round 2 of the TrojAI benchmark featured 1,104 image-classification models trained on synthetic traffic sign data. Key complexities included an increased number of classes (ranging from 5 to 25), a variety of trigger types (both polygonal shapes and Instagram-like filters), selective poisoning of source classes (affecting 1, 2, or all classes), and a diverse set of 23 model architectures. The test dataset consists of 144 models.

\textbf{Round 3.} Round 3 experimental design is identical to Round2 with the addition of Adversarial Training. Two different Adversarial Training approaches: PGD and FBF \cite{Munro2020Fast} are used. The test dataset consists of 288 models. 

\textbf{Round 4.} 
Round 4 introduces more complex triggers with increased difficulty, including multiple concurrent triggers and conditional firing. Triggers are one-to-one mappings, with up to two per model, and can have spatial, spectral, or class-based conditions. These conditions determine whether a trigger activates based on location, color, or the class it is applied to, allowing for more nuanced misclassification scenarios. This round comprised 1,008 CNN models with half (50\%) of the models have been poisoned with an embedded trigger. The test dataset consists of 288 models.

\textbf{Round 11.} This dataset builds on Round 4, featuring models with up to ~130 classes and {0, 1, 2, or 4} triggers per model, including Polygon and Instagram filter types. Triggers can have spatial, spectral, or texture conditionals, requiring specific location, color, or texture to activate misclassification. The round also introduces more spurious triggers (inactive or in clean models) to make actual triggers more targeted and specific. The training dataset consists of 288 models, the test dataset consists of 216 models, and the holdout dataset consists of 216 models.

\textbf{CIFAR-10.} Introduced in 2009, the CIFAR-10 \cite{krizhevsky2009learning} collection comprises 50,000 images for model training and 10,000 for evaluation. Every sample is a 32x32-pixel RGB image (3 channels) categorized into ten distinct groups, such as animals, vehicles, and everyday objects. This dataset is widely utilized for benchmarking classification algorithms in machine learning.

\textbf{GTSRB.} The German Traffic Sign Recognition Benchmark (GTSRB) \cite{houben2013detection}, released in 2013, focuses on identifying 43 types of road signs. With 39,209 training images and 12,630 test images, the dataset exhibits class imbalance, as certain sign categories appear more frequently than others. To standardize inputs, all images are adjusted to 32x32 pixels across three color channels, matching the dimensions of other datasets commonly used in similar research.

\textbf{ImageNet.} A widely recognized resource in computer vision, ImageNet \cite{deng2009imagenet} includes millions of labeled images spanning thousands of categories. For practical experimentation, a condensed version is often employed—here, 100 classes were chosen, each containing 500 training and 100 validation images. These high-resolution samples (224x224 pixels, three channels) enable testing models on more complex visual data compared to smaller datasets like CIFAR-10. The full ImageNet repository remains a cornerstone for training deep neural networks.

%% file: sec/Appendix/algorithm.tex


\algrenewcommand\algorithmicrequire{\textbf{Inputs:}}
\algrenewcommand\algorithmicensure{\textbf{Outputs:}}

\resizebox{\textwidth}{!}{%
  \begin{minipage}{\textwidth}
\begin{algorithm}[H]
\caption{Trigger Reconstruction and Trojan Detection via Diffusion Guidance}
\label{alg:RET}
\begin{algorithmic}[1] 
\Require 
    
    Classifier $f$ under test, pretrained diffusion model with denoising functions $\mu_{\theta}(\cdot)$ and $\Sigma_{\theta}(\cdot)$, hyperparameters $\lambda_1, \lambda_2$, candidate label sets $\mathcal{Y}^{\text{src}}$ (source) and $\mathcal{Y}^{\text{tar}}$ (target) with $y^{\text{src}} \neq y^{\text{tar}}$, number of diffusion steps $T$, and (optional) clean source data $\mathcal{X}^{\text{src}}$ for hybrid conditioning.
\Ensure Decision (Trojaned / Clean) and, if Trojaned, the corresponding trigger $\delta^{\text{tar}}_{\text{src}}$.
    
\For{each label pair $(y^{\text{src}}, y^{\text{tar}})$ with $y^{\text{src}} \neq y^{\text{tar}}$}
    \Repeat
        \State \textbf{Initialize:} Sample $x_T \sim \mathcal{N}(0, I)$.
        \For{$t=T$ \textbf{downto} $1$}
            \State Compute gradient:
            \If{clean source data $\mathcal{X}^{\text{src}}$ is provided}
                \State Compute hybrid gradient:
                \[
                g_t \leftarrow \nabla_{x_t}\log\frac{f(y^{\text{tar}} \mid \mathcal{X}^{\text{src}}\oplus x_t)}{f(y^{\text{src}} \mid \mathcal{X}^{\text{src}}\oplus x_t)}
                \]
            \Else
                \State Compute standard gradient:
                \[
                g_t \leftarrow \nabla_{x_t}\log\frac{f(y^{\text{tar}} \mid x_t)}{f(y^{\text{src}} \mid x_t)}
                \]
            \EndIf

            \State Sample uniform noise: $\eta_t \sim \lambda_1\cdot t \cdot\mathcal{U}(0,1)$.
            \State Compute modified mean:
            \[
            \tilde{\mu}_{\theta}(x_t, t, y^{\text{tar}}, y^{\text{src}})=\mu_{\theta}(x_t, t)+ \,\Sigma_{\theta}(x_t, t)\,g_t+\eta_t.
            \]
            \State Sample $x_{t-1} \sim \mathcal{N}\Bigl(\tilde{\mu}_{\theta}(x_t, t, y^{\text{tar}}, y^{\text{src}}),\,\Sigma_{\theta}(x_t, t)\Bigr)$.
        \EndFor
        \State Set the generated trigger candidate: $\delta^{\text{tar}}_{\text{src}} \leftarrow x_0$.
    \Until{$\text{softmax}[f(\delta^{\text{tar}}_{\text{src}})]_{y^{\text{tar}}} \geq 0.9$}
    \State Compute trigger strength:
\begin{equation*}
\begin{aligned}
    S(y^{\text{src}},y^{\text{tar}}) \leftarrow 
    \mathbb{E}_{x'\sim \mathcal{X}'^{\text{src}}} \Bigl[ 
    & \text{softmax}\bigl(f(x'+\delta^{\text{tar}}_{\text{src}})\bigr)_{y^{\text{tar}}}  -\text{softmax}\bigl(f(x'+\delta^{\text{tar}}_{\text{src}})\bigr)_{y^{\text{src}}} 
    \Bigr].
\end{aligned}
\end{equation*}

\EndFor
\State Identify the pair with maximum trigger strength:
\[
(y^{\text{src*}}, y^{\text{tar*}}) \leftarrow \arg\max_{(y^{\text{src}}, y^{\text{tar}})} S(y^{\text{src}},y^{\text{tar}}).
\]

\If{$S(y^{\text{src*}}, y^{\text{tar*}}) \geq \lambda_2$}
    \State \Return Trojaned, $\delta^{\text{tar*}}_{\text{src*}}$.
\Else
    \State \Return Clean.
\EndIf
\end{algorithmic}
\end{algorithm}
\end{minipage}%
}

\clearpage


%% file: sec/Fig_4.tex
\begin{figure*}[t!]
  \centering
  \includegraphics[width=\linewidth]{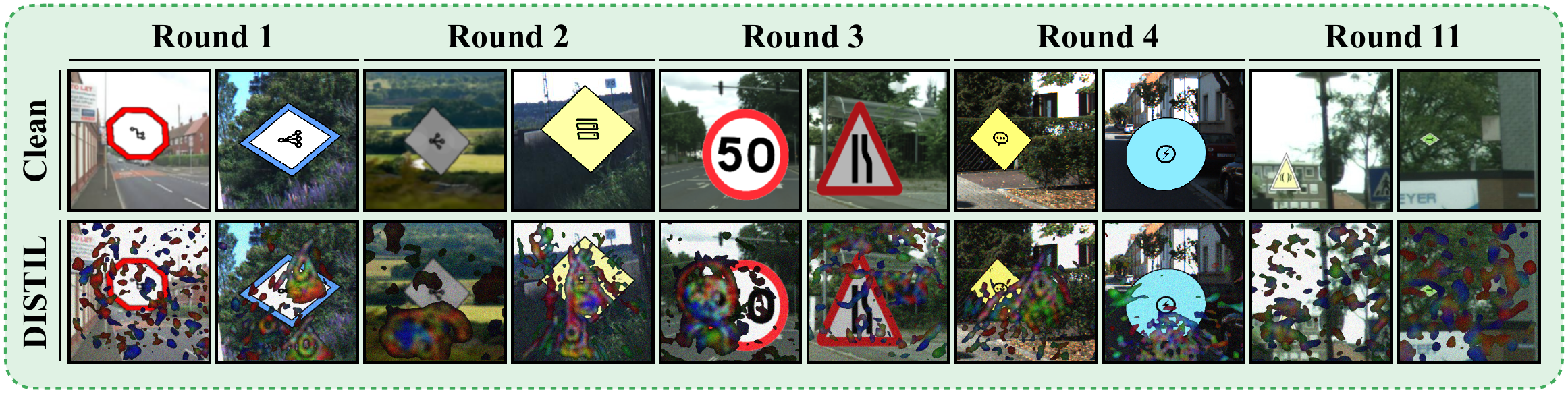}
  \caption{\textbf{DISTIL Performance on Clean Models.} This figure displays the noisy, random trigger patterns generated by DISTIL when applied to clean models across multiple rounds (Rounds 1, 2, 3, 4, and 11). As hypothesized, these triggers lack coherent structure and exhibit minimal transferability, aligning with expectations for non-Trojaned systems. The absence of consistent patterns underscores DISTIL’s specificity to compromised models, reinforcing its discriminative power in distinguishing benign systems from Trojaned ones, as validated by the proposed method and experimental results.}
  \label{fig:fig4}
\end{figure*}

%% file: sec/Fig_5.tex
\begin{figure*}[t!]
  \centering
  \includegraphics[width=\linewidth]{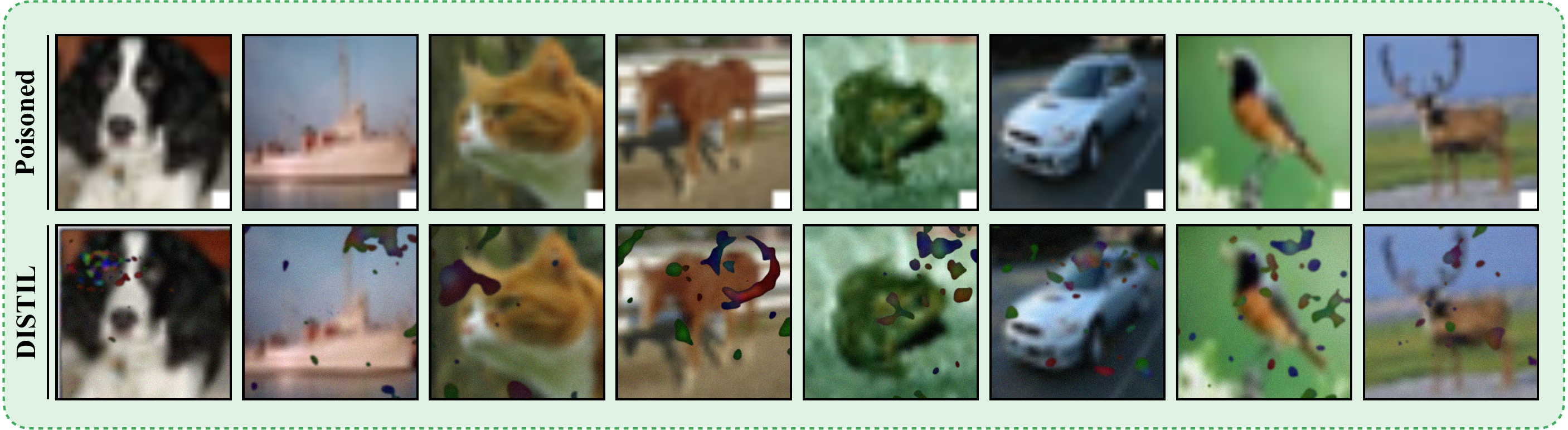}
  \caption{\textbf{DISTIL Performance on CIFAR10 with BadNets Trigger.} This figure illustrates the trigger pattern reconstruction achieved by DISTIL on the CIFAR10 dataset under a BadNets attack scenario. The extracted trigger patterns clearly reflect the characteristic artifacts of the BadNets trigger, showcasing DISTIL’s capability to accurately recover and highlight Trojan signatures even in challenging poisoning conditions.}
  \label{fig:fig5}
\end{figure*}

%% file: Tables/table_5.tex
\begin{table*}[t]
\centering
\caption{Mean $\pm$ standard deviation of AUCROC over ten independent runs for the DISTIL method on each round of the TrojAI dataset.}
\label{table:STATS}
\resizebox{0.6\textwidth}{!}{%
\begin{tabular}{l c c c c c c}
\specialrule{1.5pt}{\aboverulesep}{\belowrulesep}
    \multirow{2}{*}{\textbf{Method}} & \multicolumn{6}{c}{\textbf{Dataset}} \\
\cmidrule(lr){2-7}
    & \textbf{Round 0} & \textbf{Round 1} & \textbf{Round 2} & \textbf{Round 3} & \textbf{Round 4} & \textbf{Round 11} \\
\specialrule{1.5pt}{\aboverulesep}{\belowrulesep}
\textbf{DISTIL}  & 83.1$\pm$1.2 & 82.9$\pm$0.9 & 79.5$\pm$0.5 & 78.4$\pm$1.3 & 84.6$\pm$0.8 & 80.4$\pm$0.2 \\
\specialrule{1.5pt}{\aboverulesep}{\belowrulesep}
\end{tabular}
}
\label{table:std}
\end{table*}